%% file: main_workshop.tex
\documentclass[runningheads]{llncs}

\usepackage[T1]{fontenc}
\usepackage{amsmath}
\usepackage{multirow}
\usepackage{makecell}
\usepackage{booktabs}
\usepackage[table]{xcolor}
\usepackage{tikz}

\newcommand{\circnum}[1]{%
\tikz[baseline=(char.base)]{
\node[
    shape=circle,
    draw,
    inner sep=1pt
] (char) {\scriptsize #1};}
}

\usepackage{graphicx,verbatim}
\usepackage{amssymb}

\usepackage{pifont}% http://ctan.org/pkg/pifont
\newcommand{\cmark}{\ding{51}}%
\newcommand{\xmark}{\ding{55}}%

\usepackage[table,xcdraw]{xcolor}
\usepackage{booktabs}
\usepackage[table]{xcolor}
\usepackage{siunitx}
\usepackage{booktabs}
\usepackage[table]{xcolor}
\usepackage{makecell}
\usepackage{multirow}

\begin{document}
%
% \title{GNN-based Temporal Representation Learning of Cancer Treatment Response Trajectories: Application to pCR Prediction in Breast Cancer}

% Richard: Suggestion for a simplified and shorter title without acronyms: 
\title{Temporal Graph Representation Learning of Patient Treatment Response Trajectories}

\title{Graph Representation Learning of Longitudinal Medical Imaging Trajectories for Patient Treatment Response Prediction}

\title{Graph Representation Learning of Longitudinal Medical Imaging Trajectories for Treatment Response Prediction}

\titlerunning{Temporal Graph %Neural Network 
Representation Learning of Patient %Treatment Response 
Trajectories %:Application to Breast Cancer Response Prediction
}

\begin{comment}  %% Removed for anonymized MICCAI submission
\author{First Author\inst{1}\orcidID{0000-1111-2222-3333} \and
Second Author\inst{2,3}\orcidID{1111-2222-3333-4444} \and
Third Author\inst{3}\orcidID{2222--3333-4444-5555}}
%
\authorrunning{F. Author et al.}
\institute{Princeton University, Princeton NJ 08544, USA \and
Springer Heidelberg, Tiergartenstr. 17, 69121 Heidelberg, Germany
\email{lncs@springer.com}\\
\url{http://www.springer.com/gp/computer-science/lncs} \and
ABC Institute, Rupert-Karls-University Heidelberg, Heidelberg, Germany\\
\email{\{abc,lncs\}@uni-heidelberg.de}}

\end{comment}

% \author{Anonymized Authors}  %% Added for anonymized MICCAI submission
\author{Johannes Kiechle$^{1,2,3,7}$, Richard Osuala$^{2,4}$, Daniel M. Lang$^{3}$, \\Stefan M. Fischer$^{1,2,3,7}$, Ivana Janíčková$^{5}$, Karim Lekadir$^{4}$, \\Julia A. Schnabel$^{1,3,6,7,\dagger}$, and Jan C. Peeken$^{2,\dagger}$}  %% Added for anonymized MICCAI submission
\authorrunning{Kiechle et al.}
\institute{$^1$ Technical University of Munich, $^2$ TUM University Hospital Rechts der Isar, \\$^3$ Helmholtz Munich, $^4$ Universitat de Barcelona, $^5$ Medical University of Vienna, \\$^6$ King’s College London, $^7$ Munich Center for Machine Learning
}
  
\maketitle              % typeset the header of the contribution

\begingroup
\renewcommand{\thefootnote}{\fnsymbol{footnote}}
\footnotetext[4]{Shared senior authorship.}
\endgroup

\begin{abstract}

% Breast cancer is 
%among the most prevalent diagnosed cancers worldwide and the 
%by far 
% the most common malignancy in women, with an estimated 2.3 million new cases and approximately 670,000 deaths each year. Commonly treated with neoadjuvant chemotherapy (NACT), prior studies indicate pathological complete response (pCR) as a meaningful surrogate marker for long-term outcomes. %in high-risk breast cancer. 

In patients with breast cancer, pathological complete response (pCR) has been established as a clinically meaningful surrogate marker for long-term outcomes. While commonly treated with neoadjuvant chemotherapy (NACT), effective treatment decision-making remains challenging, as therapeutic response can vary substantially across patients, calling for predictive models capable of accurately estimating individualized treatment response. To address this, we propose an imaging-based 3D spatio-temporal framework for treatment response prediction that integrates a state-of-the-art graph neural network with relational modeling of temporal interactions across timepoints alongside three novel complementary self-supervised treatment trajectory representation learning objectives. 
%effectively combining a CNN feature extractor with a GNN-based projection head. Moreover, we introduce an asymmetric pCR loss to facilitate learning of breast cancer treatment response trajectories, which is specifically designed to appropriately account for responder patients and clinically ill-defined non-responder cases. 
Experiments across a cohort of 585 patients from the public ISPY-2 dataset demonstrate that our method substantially outperforms both vision and self-supervised learning baselines across several classification metrics. Alongside establishing a breast cancer pCR prediction benchmark, we include a principled ablation of our method and further introduce and empirically assess the impact of the available number of DCE-MRI timepoints per patient trajectory and the inclusion of inter-scan time-differences. Overall, our study substantiates the utility of clinically meaningful longitudinal medical imagaging modeling for predicting NACT-induced pCR. We will publicly share our code repository and a user-friendly PyPI library for dataset curation upon publication, effectively promoting reproducible open-source research.

% We publicly share our code repository at \url{https://github.com/anonymous} alongside a user-friendly Python package for dataset curation at \url{https://pypi.org/project/anonymous}, effectively promoting reproducible open-source research.

% Authors must provide keywords and are not allowed to remove this Keyword section.

\end{abstract}
%
%
%

%\section{Introduction and Related Work (Page 1-2)}

%\textbf{Ideas for Introduction}
%\begin{itemize}
%    \item We should mention in the introduction that some spatio-temproal learning approaches in breast cancer pCR prediction such as Jing et al.~\cite{jing2024prediction} and Janickova et al.~\cite{janivckova2025temporal} lack a simple CNN baseline which usually indicates the lower bound for subsequent spatio-temporal methodology development. We do that and therfore provide a holistic benchmark on top of our novel method for pCR prediction based on GNNs.
%    \item We should mention that in treatment response prediction scenarious it is usually the case that the non-responder class is ill-defined, which means that, in contrast to reponder patients, the class of non-responders may include several sub-classes such as "progressors", "partial-responders" and patients with "stable" disease, which are semantically different and needs to be properly addressed during training --> we do that by means of our "asymmetric" loss which is specifically targeted for responder patients, as we can be sure that the label for all patients in this group is correct.
%\end{itemize}

\begin{figure}
\includegraphics[width=\textwidth]{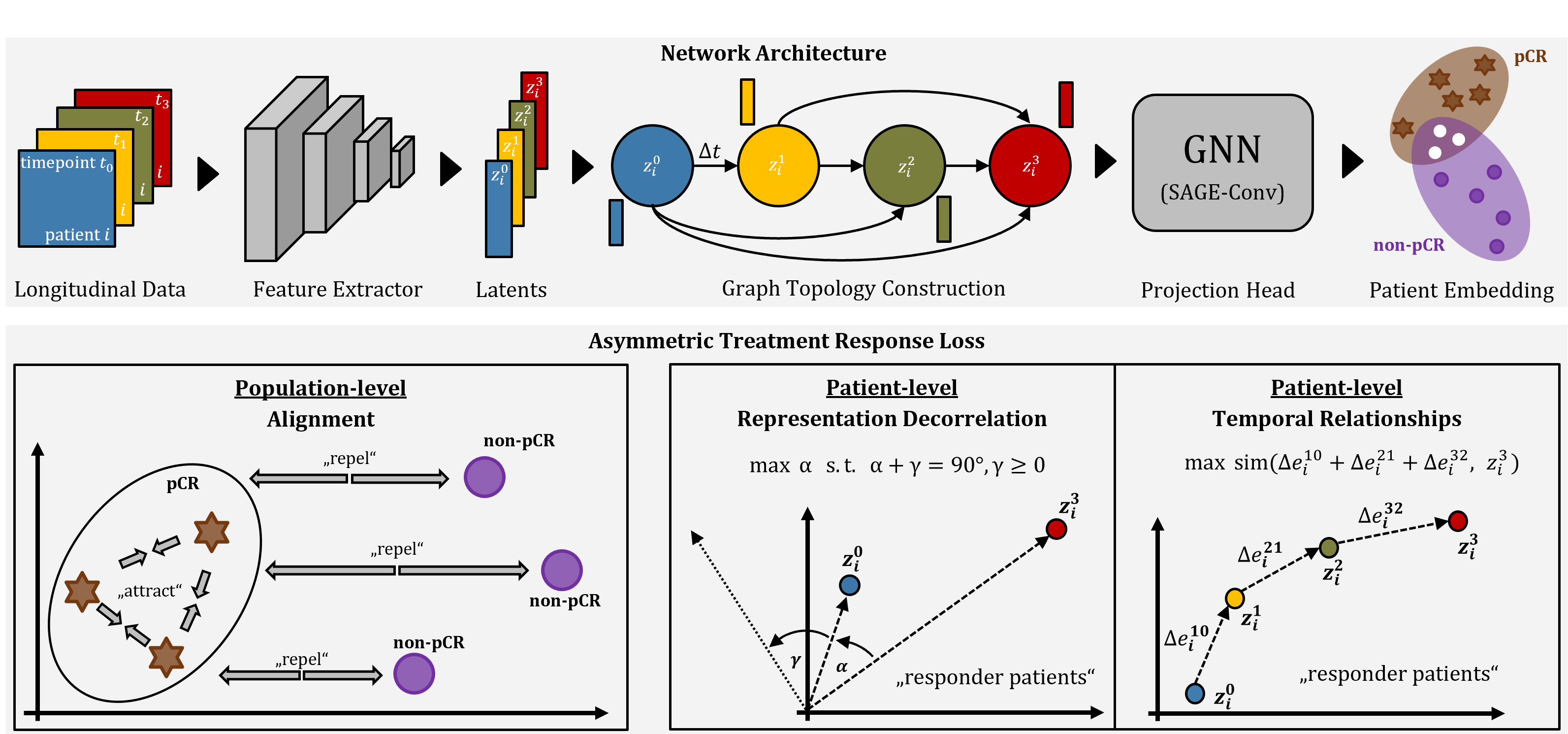}
% \caption{Methodology: Longitudinal latent features extracted from patient images are organized as a directed acyclic graph, before being aggregated by a GNN projection head to produce a compact time-aware patient embedding. Representations are learned via an asymmetric treatment response objective combining (a) population-level alignment, (b) patient-level representation decorrelation, and (c) temporal consistency loss functions, together separating latents of responders from non-responding patients while preserving longitudinal structure.} 
\caption{Method overview. Longitudinal image-derived latent features are represented as a directed acyclic graph and aggregated through a graph neural network (GNN) projection head to obtain a compact patient-level embedding. Representation learning is guided by an asymmetric response-aware objective that combines (a) population-level alignment, (b) patient-level representation decorrelation, and (c) temporal consistency losses. The overall objective encourages discriminative separation between responders and non-responders while preserving the temporal structure of disease evolution.
}
\label{fig:method}
\end{figure}

\section{Introduction}

%%%%%%%% What is the problem and why important?
%With an estimated 2.3 million women diagnosed worldwide in 2022, breast cancer accounts for approximately 25\% of all new cancer cases and 16\% of cancer-related deaths in women \cite{WHO2026}. 
With an estimated 2.3 million women diagnosed with breast cancer annually~\cite{WHO2026}, optimizing treatment remains a clinical priority where neoadjuvant chemotherapy (NACT) crucially enables early response assessment to guide therapy adaptation and surgical planning~\cite{li2020predicting}. Pathologic complete response (pCR), i.e. the absence of residual invasive disease in the breast and axillary lymph nodes at surgery following NACT~\cite{li2020predicting}, is commonly used to estimate NACT efficacy while being associated with improved long-term outcomes and event-free survival~\cite{spring2020pathologic}.

%, including event-free and overall survival, particularly in aggressive subtypes such as triple-negative and HER2-positive breast cancer \cite{spring2020pathologic}. 
%
%%%%%%%% Now that we know the problem What can help to solve it and why has it been failing?
Reliable early prediction of pCR holds the promise of mitigating treatment-related toxicity and crucially supports informed clinical decision-making, including therapeutic escalation or de-escalation strategies. Moreover, early identification of non-responders may enable timely treatment modification and optimization of surgical intervention.
% can enable reduce of unnecessary toxicity and bears potential to support surgical de-escalation or expedited treatment modification and surgical intervention in early-identified non-responders. 
% With dynamic contrast-enhanced magnetic resonance imaging (DCE-MRI) playing a central role in breast cancer prognosis and NACT therapy monitoring \cite{Hylton2016}, DCE-MRI–based pCR prediction has attracted substantial interest, however, many studies only use single-modality pre-treatment input data, lack external validation or rely on single-split evaluations, limiting confidence in generalizability \cite{Hylton2016,li2020predicting,peng2022pretreatment,khan2022deep,syed2023machine,jing2024prediction,li2024breast,krasniqi2025multimodal,janivckova2025temporal}.
Dynamic contrast-enhanced magnetic resonance imaging (DCE-MRI) plays an essential role in screening and assessment of response to NACT~\cite{Hylton2016}. Accordingly, DCE-MRI–based methods for pCR prediction have attracted substantial attention. Many existing studies, however, are limited by the use of single-modality pre-treatment data, absence of rigorous baseline examinations, or reliance on single train-test split, thereby constraining confidence in generalizability of reported performances \cite{Hylton2016,li2020predicting,peng2022pretreatment,khan2022deep,syed2023machine,jing2024prediction,li2024breast,krasniqi2025multimodal,janivckova2025temporal}.
%DCE-MRI based pCR prediction has been attracting increasing research attention, despite still often lacking external validation or overestimating generalizability due to single data source evaluation on single random splits
%
%\cite{Hylton2016,li2020predicting,peng2022pretreatment,khan2022deep,syed2023machine,jing2024prediction,li2024breast,krasniqi2025multimodal,janivckova2025temporal}. %we could remove some references in case the page limit is tough to reach.
% Using images from a single timepoint (e.g. pre-treatment) provides limited insight into subtle therapy-induced changes \cite{jing2024prediction}. Additional longitudinal DCE-MRI acquired during NACT \cite{li2022spy,newitt2025acrin} can thus allow pCR prediction models to capture dynamic tumor evolution signaling individual treatment trajectories.
While imaging data from a single time point (e.g. pre-treatment) insufficiently captures subtle therapy-induced alterations \cite{jing2024prediction}, the integration of longitudinal DCE-MRI acquired over the course of NACT \cite{li2022spy,newitt2025acrin} allow pCR prediction models to capture dynamic tumor evolution, signaling individual treatment trajectories.

%%%%%%%% Related work
% Nonetheless, few studies have leveraged longitudinal DCE-MRI \cite{janivckova2025temporal}, while further containing limitations in learning complex temporal representations, e.g. by solely relying on hand-crafted features \cite{syed2023machine} or simple concatenation of timepoints \cite{krasniqi2025multimodal}. For instance, \cite{comes2021early} used multiple DCE-MRI timepoints but adopted a fixed CNN pretrained on natural images to extract latent representations, while \cite{jing2024prediction} trained an LSTM layer to learn multi-timepoint temporal DCE-MRI relationships albeit only based on the supervised binary pCR classification objective without dedicated representation learning. Similarly, \cite{zhang2024m2fusion} introduced a temporal fusion module, with attention-based and sequence modeling mechanisms, to aggregate features across longitudinal MRI timepoints, however, as well solely optimized via a supervised pCR classification objective. 
Nonetheless, only a limited number of studies have utilized longitudinal DCE-MRI data~\cite{janivckova2025temporal}, and existing approaches remain constrained in their ability to learn complex temporal representations. Several methods rely on hand-crafted features \cite{syed2023machine} or simple concatenation of multiple time points \cite{krasniqi2025multimodal}, which may inadequately capture longitudinal tumor dynamics. For example, \cite{comes2021early} incorporated multiple DCE-MRI time points but employed a fixed CNN pretrained on natural images for feature extraction, without adapting representations to the temporal treatment setting. Similarly, \cite{jing2024prediction} applied an LSTM to model inter-timepoint dependencies, however, training was restricted to a supervised binary pCR classification objective, without dedicated temporal representation learning. Likewise, Zang et al. \cite{zhang2024m2fusion} proposed a temporal fusion module integrating attention-based and sequence modeling mechanisms to aggregate longitudinal MRI features, yet optimization was again limited to supervised pCR classification.

Recently, a purely imaging-based 2D longitudinal representation learning approach \cite{janivckova2025temporal} demonstrated promising results but evaluated performance on a single dataset split, likely overestimating generalization performance. While pointing towards the needs of (i) deeper multi-split validation  and (ii) CNN-based feature extractor and self-supervised baseline comparisons \cite{janivckova2025temporal,jing2024prediction,santeramo2018longitudinal}, these previous studies signaled potential capabilities of temporal representation learning of disease trajectories. Although these studies suggest potential, its downstream impact, specifically for breast cancer pCR prediction, including varying numbers of DCE-MRI time points remains insufficiently explored \cite{janivckova2025temporal,kachele2025tackling}. We aim to bridge this gap by introducing a time-aware graph neural network with novel temporal representation learning objectives, alongside a principled pCR prediction framework and benchmark evaluation. Our key contributions are as follows:
\begin{enumerate}
\item We propose a novel time-aware graph neural network that explicitly encodes structured treatment response trajectories within its graph topology.
% \item We introduce three novel representation learning objectives for latent space alignment with longitudinal patient trajectory.
\item We introduce three novel representation learning objectives that enforce trajectory and response-consistent latent representations.
\item We establish a comprehensive imaging-based benchmark for breast cancer pCR prediction that evaluates different methods, available imaging timepoints, and the inclusion of inter-scan temporal differences.

% including a per-method impact assessment of the available number of DCE-MRI timepoints and the inclusion of inter-scan time-difference. %across supervised and self-supervised methods.
\end{enumerate}

\section{Methods and Materials}

\subsection{Network Architecture}
\noindent \textbf{Visual Feature Extraction} Given a 3D input volume, representing a patient~$i$ at timepoint~$t$, the longitudinal MRI set is described as $\{x^{t}_{i}\}_{t=1}^{T}$, where $x_i \in \mathbb{R}^{C \times D\times H \times W}$. Each input volume is independently processed by a ResNet~\cite{he2016deep} backbone to extract corresponding feature maps, which are subsequently passed through a global average pooling layer, to obtain a compact embedding vector representing an 3D input volume of patient~$i$ at timepoint~$t$, denoted as $z_i^t \in \mathbb{R}^{D}$. \newline

\noindent \textbf{Graph Topology Construction} The set of image embeddings $\{z^{t}_{i}\}_{t=1}^{T}$ representing a patient $i$ across $T$ timepoints is subsequently translated into a directed graph, where each node corresponds to one embedding $z_i^t \in \mathbb{R}^{D}$ and edges encode pairwise relationships. Therein, nodes are indexed according to their temporal ordering. For every pair of nodes $(u,v)$ with $u < v$, we add a directed edge from node $u$ to node $v$, with no reverse edges, resulting in a set of edges $\mathcal{E} = \{(u,v) \;| \; 0 \le u < v < N\}$. This yields a directed acyclic graph, which ensures that each node can propagate information to all subsequent nodes while preserving temporal dependencies. The choice of this topology ensures pairwise node connectivity under a strict ordering constraint, enabling relational modeling while maintaining a well-defined directional flow of temporal information. \newline

\noindent \textbf{Graph Feature Representation Learning} Let $\mathcal{G}=(\mathcal{V},\mathcal{E})$ denote the aforementioned directed acyclic graph, where $\mathcal{V}$ refers the set of nodes which are connected by a set of edges $\mathcal{E} \subseteq \mathcal{V} \times \mathcal{V}$. Given a feature vector $z_i^t \in \mathbb{R}^D$ attached to the corresponding node, following their temporal ordering, we perform graph feature representation learning over the nodes using GraphSAGE~\cite{hamilton2017inductive}, which follows the scheme of message passing. Specifically, nodes are updated by neighborhood feature aggregation, following the underlying graph structure~\cite{gilmer2017neural}. 
% \begin{equation} \label{eq:GraphSAGE}
%     \mathbf{h}_i = \phi \left( \mathbf{W} \left[ \mathbf{x}_i \: \|  \:\bigoplus_{j \in \mathcal{N}_i} \left( \left\{\mathbf{e}_{ji} \cdot \mathbf{x}_j, \: \forall j \in \mathcal{N}(i) \right\} \right) \right] \right)\enspace,
% \end{equation}
% Maybe: Longitudinal Responder Trajectory Losses (or as section title: "Longitudinal Responder Trajectory Representation Learning"
\subsection{Asymmetric Treatment Response Loss}
\noindent \textbf{Population-level Alignment} We propose a population-level alignment objective that structures the embedding space according to the response status of a patient (Figure~\ref{fig:method}, bottom left). Let $z_i \in \mathbb{R}^D$ denote the embedding of a patient $i$, our goal is twofold: first, encourage similarity among responder patients (i.e. attract embeddings) and second, discourage similarity between responders and non-responders (i.e. repel embeddings), based on Cosine similarity, outlined as

\begin{equation}
\mathcal{L}_{\text{align}} =
\underbrace{
1
-
\frac{1}{|\mathcal{R}|}
\sum_{i \in \mathcal{R}}
\mathrm{sim}(z_i, z_{\pi(i)})
}_{\text{\circnum{1} attract}}
+
\underbrace{
\frac{1}{|\mathcal{N}|}
\sum_{(i,j) \in (\mathcal{N}, \mathcal{R})}
\mathrm{sim}(z_i, z_j).
}_{\text{\circnum{2} repel}}
\end{equation}
$\textbf{\circnum{1} attract}\quad$ To promote a shared latent space among responder patients $\mathcal{R}$, we randomly permute the embeddings within the responder group and align each responder with another responder. Formally, let $\pi$ be a random permutation over indices in $\mathcal{R}$, the responder attract term aims at increasing the pairwise similarity within the responder population, encouraging embeddings to occupy a compact region in latent space. Moreover, as the permutation $\pi$ changes across iterations, the loss approximates a stochastic estimate of responder coherence rather than enforcing fixed pairwise matches within this subgroup. 

$\textbf{\circnum{2} repel}\quad$ To promote discriminative separation between responders $\mathcal{R}$ and non-responders $\mathcal{N}$, we penalize similarity across these groups. Minimizing the repel term, pushes embeddings of the two subpopulations apart, thereby increasing inter-class separation. Furthermore, we intentionally omit the attract term for non-responders, as their heterogeneous clinical outcomes such as stable disease, progressive disease, or partial response are associated with distinct underlying imaging patterns, making a shared latent representation ill-posed.\newline

% We do not apply the attract term to non-responders, as this cohort comprises clinically heterogeneous outcomes, including stable disease, progressive disease, and partial response. Consequently, enforcing a shared latent representation for this group would be inappropriate, as these outcomes are characterized by distinct underlying imaging patterns.
% We intentionally do not enforce the attract term among non-responders, as this heterogeneous group aggregates clinically distinct patient outcomes such as stable, progressive, and partial response (i.e. distinct image semantics) making a shared latent representation ill-posed. \newline

\noindent \textbf{Patient-level Representation Decorrelation} We propose a patient-level representation decorrelation objective, where the key idea is that embeddings from different timepoints of the same patient should not collapse to redundant representations, but rather encode complementary aspects of disease evolution. To this end, we penalize similarity between adjacent timepoint embeddings of responder patients $\mathcal{R}$ using Cosine similarity, formally defined as
\begin{equation}
\mathcal{L}_{\text{decorrelate}}
=
\frac{1}{|\mathcal{R}|}
\sum_{i \in \mathcal{R}}
\frac{1}{T-1}
\sum_{t=0}^{T-2}
\mathrm{sim}\!\left(z_i^{(t)}, z_i^{(t+1)}\right).
\end{equation}
Minimizing the objective above encourages embeddings from different timepoints of the same patient to become decorrelated. As a consequence, each temporal segment $z_i^{(t)}$ of a patient trajectory is promoted to capture distinct, non-redundant factors of variation relevant for treatment response.\newline 

\noindent \textbf{Patient-level Temporal Relationships} We propose a patient-level temporal relationship loss, with the key idea of modeling how latent embeddings $\{z^{t}_{i}\}_{t=0}^{T-1}$ of patient $i$ evolve over time. While the decorrelation term enforces complementary information across timepoints, we now constrain the latent trajectory to follow a consistent additive progression $\tilde{z}_i$, formally defined as
%
% \begin{equation}
%     \mathcal{L}_{\text{temporal}} = \frac{1}{\mathcal{R}} \sum_{i \in \mathcal{R}} 1 - \text{sim}(\tilde{z}_i, z_i^{T-1}) \, \text{with} \; \tilde{z}_i =  \sum_{t=0}^{T-2} (z_i^t-z_i^{t+1})
% \end{equation}
%
\begin{equation}
\mathcal{L}_{\text{temporal}}
=
\frac{1}{|\mathcal{R}|}
\sum_{i \in \mathcal{R}}
\Bigl[
1 - \mathrm{sim}\!\left(\tilde{z}_i, z_i^{(T)}\right)
\Bigr] \;\;\; \text{where}
\quad
\tilde{z}_i
=
\sum_{t=0}^{T-2}
\left(z_i^{(t+1)} - z_i^{(t)}\right).
\end{equation}

Minimizing the objective above encourages temporal consistency. The representation at the final timepoint $T$ must be predictable from the sequence of intermediate latent transitions. Intuitively, the embeddings are encouraged to follow a coherent temporal trajectory rather than independent snapshots, thereby regularizing the model to encode meaningful disease progression dynamics. \newline 

\noindent \textbf{Loss Function} The overall \emph{Asymmetric Treatment Response Loss} applies the introduced objectives asymmetrically to responders $\mathcal{R}$ and non-responders $\mathcal{N}$:
\[
\mathcal{L}_{\mathrm{total}}(i)
=
\mathbf{1}_{\{i \in \mathcal{R}\}}
\big(
\mathcal{L}_{\mathrm{align}}
+
\mathcal{L}_{\mathrm{decorrelate}}
+
\mathcal{L}_{\mathrm{temporal}}
\big)
+
\mathbf{1}_{\{i \in \mathcal{N}\}}
\mathcal{L}_{\mathrm{repel}}
\]

% \vspace{0.2cm} \textbf{Loss Function} The overall \emph{Asymmetric Treatment Response Loss} applies different objectives to responders and non-responders. For responder patients $\mathcal{R}$, alignment, disentanglement, and temporal consistency terms are optimized, whereas non-responders $\mathcal{N}$ are instead repelled from the responder representation space:

% \begin{equation}
% \mathcal{L}_{\mathrm{total}} =
% \begin{cases}
% \mathcal{L}_{\mathrm{align}}
% + \mathcal{L}_{\mathrm{disentangle}}
% + \mathcal{L}_{\mathrm{temporal}}
% & \text{if } i \in \mathcal{R}, \\[4pt]
% \mathcal{L}_{\mathrm{repel}}
% & \text{if } i \in \mathcal{N}.
% \end{cases}
% \end{equation}

\section{Experiments and Results}

\noindent \textbf{Data and Preprocessing} In this work we use the publicly available \mbox{ISPY-2} dataset~\cite{li2022spy,newitt2025acrin} and select 585 patients (204 responders) with complete DCE-MRI acquisitions at four neoadjuvant chemotherapy (NACT) timepoints, comprising one pre-treatment scan and three consecutive examinations acquired during therapy prior to surgery. Data preprocessing follows Jing et al.~\cite{jing2024prediction}, resulting in a longitudinal sequence of four images per patient, each represented as a tensor of shape $(3,64,64,64)$. The three channels encode early contrast enhancement (phase 1 minus phase 0), late contrast enhancement (phase 4 minus phase 0), and functional tumor volume (FTV), which captures metabolically active tumor regions defined via pharmacokinetic thresholds on DCE-MRI~\cite{newitt2014real}. To ensure reproducibility and enable fair comparison, we release a library at the Python Package Index (PyPI) implementing the full data pipeline upon publication.

\noindent \textbf{Implementation Details} All models are trained and evaluated using stratified 5-fold cross-validation (3/1/1 train/val/test split per fold). Optimization is performed with SGD (learning rate $10^{-2}$, and batch size 16 with gradient accumulation over 4 batches) for 100 epochs using a cosine annealing scheduler. The model with the highest validation AUC is selected for testing. To ensure fair comparison, all methods share identical data splits, inputs, and the same ResNet18 feature encoder implementation~\cite{cardoso2022monaiopensourceframeworkdeep}, such that performance differences reflect only the respective spatio-temporal learning strategies. DINOv3~\cite{simeoni2025dinov3} is included as a strong training-free foundational model baseline, using the 2D axial volume slice containing the largest metabolically active tumor region (based on the FTV channel). All vision baselines employ the same three-layer MLP classification head, with the CNN+LSTM approach additionally incorporating an LSTM layer for sequential modeling~\cite{jing2024prediction}. Self-supervised methods are evaluated using a support vector classifier. Experiments are conducted on an NVIDIA RTX A6000 Ada GPU. Further details can be found in our publicly available code repository.

\begin{table}[t]
\centering
\caption{Quantitative comparison of GNN-pCR with vision and self-supervised learning (SSL) baselines, with an ablation study of its key components. Performance is reported as the mean value over 5-folds, using balanced accuracy (bACC), Area Under the Receiver Operating Characteristic Curve (AUC) and Matthews correlation coefficient (MCC). \textbf{Best in bold}, \underline{second best underlined} and $\uparrow$ means higher is better. Wilcoxon signed-rank test: $^{(*)}$refers to p-values $\le$ 0.05, $^{(\dagger)}$refers to p-values $\le$ 0.0625.}
\label{table1}
\setlength{\tabcolsep}{6pt}
\renewcommand{\arraystretch}{1.2}

\input{tables/table1}
\end{table}

\subsection{Empirical Results and Experimental Analysis}

% In the following we provide the results of our experimental analysis, wherein we evaluate all methods based on Area Under the Receiver Operating Characteristic Curve (AUC), balanced accuracy (bACC) and Matthews correlation coefficient (MCC). 

% \subsubsection{Pathologic Complete Response Prediction} 
Table~\ref{table1} presents a quantitative comparison between our proposed GNN-pCR framework and several vision and self-supervised learning (SSL) baselines for the task of binary breast cancer pCR prediction. 
We evaluated~\cite{janivckova2025temporal} on the reference 2D data using 5-fold cross-validation and, in line with the authors~\cite{cirmuw_temporal_representation_learning_2026}, observed substantial performance variability, achieving an AUC of $0.5610$. To ensure fair evauluation, we extended the method to 3D using our unified encoder, denoted as $\text{3D-L}_{\text{ART}}$.
% We evaluated~\cite{janivckova2025temporal} using the reference 2D data across 5-folds. In line with the observations by the authors, we observe a substantial variation in performance, and only obtained $0.5610$ (AUC). To ensure fairness, we extend the method to 3D with our unified encoder and refer to it as $\text{3D-L}_{\text{ART}}$. 
Overall, our GNN-pCR consistently achieves the best performance across all evaluation metrics. In particular, compared to the strongest respective baseline model, our approach improves bACC from $0.6520$ \cite{janivckova2025temporal} to $0.6844$,
%(Janickova et al.)
AUC from $0.6951$ (best CNN baseline) to $0.7203$, and Matthews correlation coefficient (MCC) from $0.2922$ \cite{janivckova2025temporal} to $0.3561$.
%(Janickova et al.)
These results suggest that structured modeling of longitudinal relationships as proposed in our GNN-pCR approach, successfully translates into consistent improvements across all evaluation criteria and models. The performance improvement is even larger when compared to the SSL approach of 
%Kaczmarek et al.
~\cite{kaczmarek2025ssl}, the sequence modeling strategy based on CNN+LSTM~\cite{jing2024prediction} and the recent transformer vision 
foundation model DINOv3~\cite{simeoni2025dinov3}. %known as particularly strong %representation 
%feature encoder.

\noindent \textbf{Ablation Study} We conduct an ablation study by systematically removing individual loss terms and the GNN module from the full model while keeping all other components untouched.
% To assess the contribution of the individual components of our proposed framework, we conduct a comprehensive ablation study. Starting from the full model, we systematically remove individual loss terms and the GNN module, while keeping other factors untouched. 
Results can be found in Table~\ref{table1}. Removing the alignment loss leads to a consistent performance drop across all metrics, with a substantial decline in e.g., bACC from $0.6844$ to $0.6162$. This finding is consistent with the objective of enhancing class separability between responders and non-responders, indicating that the alignment loss constitutes a key factor in shaping discriminative pCR representations. Similarly, removing the temporal loss ($0.6844 \rightarrow 0.6589$) or decorrelation loss ($0.6844 \rightarrow 0.6350$) leads to considerable performance declines across all metrics. These findings suggest that both encouraging complementary features (via decorrelation loss) and enforcing temporal consistency (via temporal loss) are essential to capture relevant longitudinal changes in responder patients. Finally, replacing the GNN with a linear prediction head also reduces performance ($0.6844 \rightarrow 0.6610$), underscoring that relational modeling of temporal interactions across timepoints provides benefits beyond those achievable with dense feature connections alone.

\begin{table}[t]
\centering
\caption{Quantitative comparison of methods for (i) early response prediction and (ii) impact of inter-scan time differences. Performance is reported as the mean value over 5-folds, using balanced accuracy (bACC), for consecutive timepoints $t_0 \rightarrow t_1$, $t_0 \rightarrow t_2$ and $t_0 \rightarrow t_3$. \textbf{Best in bold}, \underline{second best underlined} and $\uparrow$ means higher is better.}

\label{table2}
\setlength{\tabcolsep}{6pt}
\renewcommand{\arraystretch}{1.2}

\input{tables/table2}
\end{table}

\noindent \textbf{Early Response Prediction} To assess the capability of the models to predict therapy response at early stages, we restrict the available longitudinal information to partial time series and evaluate performance for consecutive time intervals ($t_0 \rightarrow t_1$, $t_0 \rightarrow t_2$, and $t_0 \rightarrow t_3$). Results are summarized in Table~\ref{table2}. Across all temporal settings, our proposed GNN-pCR consistently achieves the strongest performance. Notably, the superiority of the full time-series model ($t_0 \rightarrow t_3$) translates to earlier prediction scenarios. When only two timepoints are available ($t_0 \rightarrow t_1$), GNN-pCR attains the highest bACC (0.5937), clearly outperforming CNN-based and recurrent baselines. The margin over the second-best SSL method ($\text{3D-L}_{\text{ART}}$) is particularly pronounced in this most challenging early setting, highlighting the benefit of our novel learning objectives combined with explicit modeling of relational structure under limited temporal data. We further analyze the impact of explicitly incorporating intra-patient time differences across scans as an additional feature. The effect is method-dependent. The CNN+LSTM~\cite{santeramo2018longitudinal} benefits most at the full time-series ($t_0 \rightarrow t_3$), improving from $0.6107$ to $0.6513$, although starting from a comparatively low score. In contrast, CNN, $\text{3D-L}_{\text{ART}}$ and GNN-pCR exhibit only moderate changes when adding time-gap information, which we attribute to the systematic design of the ISPY-2 study, which shows near homogeneous time difference distribution across patients and %within 
their respective  scans~\cite{li2022spy,newitt2025acrin}. 

% While the best overall performance is achieved by our GNN-pCR at the full timeseries, the performance successfully translates into limited timepoint scenarios at three ($t_0 \rightarrow t_2$) and two ($t_0 \rightarrow t_1$), outperforming all baseline methods. Moreover, within the given setting, we also evaluate the impact of incorporating intra-patient inter-scan time differences. At the full time-series ($t_0 \rightarrow t_3$) the CNN+LSTM~\cite{santeramo2018longitudinal} benefits most, however starting at the lowest initial performance. Other methods, including CNN, Janickova et al. and our GNN-pCR only show a moderate improvment, by appending the time gap between scans as additional feature to the learned representations.    

% While the performance improvement using inter-scan time differences is moderate across across CNN, Janickova et al. and our method, we have to acknowledge that these methods incorporate time difference as feature append. Longitudinal MRI scans acquired during neoadjuvant chemotherapy (NACT) are typically obtained at non-uniform time intervals, which may reflect both treatment dynamics and clinical scheduling constraints. Ignoring these inter-scan time differences implicitly assumes homogeneous temporal progression, potentially limiting the model’s ability to capture meaningful therapy-induced changes. We therefore investigate the impact of explicitly incorporating inter-scan time differences on downstream prediction of pathological complete response (pCR). 

\section{Discussion and Conclusion}

We have introduced a novel self-supervised framework for learning imaging-based patient treatment trajectories by combining a time-aware graph neural network with relational temporal modeling alongside three complementary trajectory-level objectives. The underlying graph structure explicitly encodes dependencies across timepoints, enabling structured representation learning aligned with clinically meaningful disease progression. Our experimental evaluations demonstrate consistent improvements over both supervised vision baselines and existing self-supervised approaches across multiple classification metrics. Ablation studies confirm the contribution of each novel loss component and highlight the added value of relational modeling via the GNN. Notably, early response prediction results suggest that the learned representations capture clinically relevant longitudinal characteristics. Incorporating inter-scan time differences, although holding promise, yielded only moderate gains, likely due to the structured study design of ISPY-2 and the resulting less pronounced variability in acquisition intervals across patients and scans. Even though our evaluations are based on breast cancer pCR prediction, our proposed framework is designed as a general approach for longitudinal SSL. 
To this end, future work will evaluate the proposed framework across a broader range of longitudinal clinical prediction and monitoring tasks, potentially in combination with complementary clinical biomarkers.

\bibliographystyle{unsrt}

\bibliography{mybibliography}

\end{document}

%% file: tables/table1.tex
{\fontsize{8}{9.6}\selectfont
\begin{tabular}{l|c|
                c
                c
                c}
\toprule
\textbf{Method} & \textbf{Backbone} &
\textbf{bACC $\uparrow$} & \textbf{AUC $\uparrow$} & \textbf{MCC $\uparrow$} \\
\midrule

\multicolumn{5}{l}{\textit{\underline{Vision baselines}}} \\

CNN                                     & ResNet18 & 0.6466$^{(\dagger)}$ {\tiny$\pm$ 0.03} & \underline{0.6951}$^{(\dagger)}$ {\tiny$\pm$ 0.03} & 0.2864$^{(\dagger)}$ {\tiny$\pm$ 0.06} \\
CNN + LSTM\textsuperscript{\cite{jing2024prediction}}    
                                        & ResNet18 & 0.6107$^{(\dagger)}$ {\tiny$\pm$ 0.05} & 0.6825 {\tiny$\pm$ 0.06} & 0.2278$^{(\dagger)}$ {\tiny$\pm$ 0.10} \\
DINOv3\textsuperscript{\cite{simeoni2025dinov3}}         
                                        & ViT-L/16 & 0.5628$^{(*)}$ {\tiny$\pm$ 0.04} & 0.6020$^{(*)}$ {\tiny$\pm$ 0.03} & 0.1316$^{(*)}$ {\tiny$\pm$ 0.09} \\
\midrule
\addlinespace

\multicolumn{5}{l}{\textit{\underline{SSL baselines}}} \\
$\text{3D-L}_{\text{ART}}$\textsuperscript{\cite{janivckova2025temporal}}  
                                        & ResNet18 & \underline{0.6520}$^{(\dagger)}$ {\tiny$\pm$ 0.03} & 0.6857 {\tiny$\pm$ 0.02} & \underline{0.2922}$^{(*)}$ {\tiny$\pm$ 0.06} \\
SSL-TOPC\textsuperscript{\cite{kaczmarek2025ssl}}        
                                        & ResNet18 & 0.6224$^{(*)}$ {\tiny$\pm$ 0.02} & 0.6775 {\tiny$\pm$ 0.04} & 0.2434$^{(*)}$ {\tiny$\pm$ 0.03} \\
\midrule
\addlinespace

\rowcolor{gray!10}
\textbf{GNN-pCR (ours)}    
                                        & ResNet18 & \textbf{0.6844} {\tiny$\pm$ 0.01} & \textbf{0.7203} {\tiny$\pm$ 0.02} & \textbf{0.3561} {\tiny$\pm$ 0.02} \\
\addlinespace
\midrule
\midrule

\multicolumn{5}{l}{\textit{\underline{Ablation study}}} \\
w/o alignment loss         & ResNet18 & 0.6162$^{(*)}$ {\tiny$\pm$ 0.03} & 0.6936 {\tiny$\pm$ 0.04} & 0.2490$^{(*)}$ {\tiny$\pm$ 0.06} \\
w/o temporal loss          & ResNet18 & 0.6589$^{(*)}$ {\tiny$\pm$ 0.03} & 0.7025 {\tiny$\pm$ 0.06} & 0.3047$^{(*)}$ {\tiny$\pm$ 0.05} \\
w/o decorrelation loss   & ResNet18 & 0.6350$^{(*)}$ {\tiny$\pm$ 0.01} & 0.6594$^{(*)}$ {\tiny$\pm$ 0.03} & 0.2684$^{(*)}$ {\tiny$\pm$ 0.03} \\
w/o GNN                    & ResNet18 & 0.6610 {\tiny$\pm$ 0.02} & 0.6895 {\tiny$\pm$ 0.04} & 0.3077$^{(\dagger)}$ {\tiny$\pm$ 0.04} \\

\bottomrule
\end{tabular}
}

%% file: tables/table2.tex
{\fontsize{8}{9.6}\selectfont
\begin{tabular}{l| c|
                c
                c
                c}
\toprule
\multirow{2}{*}{\textbf{Method}} &
\multirow{2}{*}{\makecell{\textbf{Inter-Scan}\\\textbf{Time Diff.}}} &
\multicolumn{3}{c}{\textbf{bACC $\uparrow$} } \\
\cmidrule(lr){3-5}
& & {$t_0 \rightarrow t_1$} & {$t_0 \rightarrow t_2$} & {$t_0 \rightarrow t_3$} \\
\midrule

\multirow{2}{*}{CNN}
  & \xmark & 0.5042 {\tiny $\pm$ 0.03} & 0.6508 {\tiny $\pm$ 0.03} & 0.6466 {\tiny $\pm$ 0.03} \\
  & \cmark & 0.5059 {\tiny $\pm$ 0.02} & 0.6479 {\tiny $\pm$ 0.03} & 0.6447 {\tiny $\pm$ 0.02} \\
\addlinespace
\midrule

\multirow{2}{*}{CNN + LSTM\textsuperscript{\cite{jing2024prediction,santeramo2018longitudinal}}}
  & \xmark & 0.5237 {\tiny $\pm$ 0.05} & 0.5721 {\tiny $\pm$ 0.03} & 0.6107 {\tiny $\pm$ 0.05} \\
  & \cmark & 0.5163 {\tiny $\pm$ 0.04} & 0.5969 {\tiny $\pm$ 0.03} & 0.6513 {\tiny $\pm$ 0.02} \\
\addlinespace
\midrule

\multirow{2}{*}{$\text{3D-L}_{\text{ART}}$\textsuperscript{\cite{janivckova2025temporal}}}
  & \xmark & 0.5586 {\tiny $\pm$ 0.05} & 0.6631 {\tiny $\pm$ 0.03} & 0.6520 {\tiny $\pm$ 0.03} \\
  & \cmark & 0.5369 {\tiny $\pm$ 0.06} & 0.6391 {\tiny $\pm$ 0.05} & 0.6616 {\tiny $\pm$ 0.02} \\
\addlinespace
\midrule

\rowcolor{gray!10}
  & \xmark & \underline{0.5820} {\tiny $\pm$ 0.04} & \textbf{0.6823} {\tiny $\pm$ 0.01} & \underline{0.6844} {\tiny $\pm$ 0.01} \\
\rowcolor{gray!10}
\multirow{-2}{*}{\textbf{GNN-pCR (ours)}}
  & \cmark & \textbf{0.5937} {\tiny $\pm$ 0.06} & \underline{0.6689} {\tiny $\pm$ 0.04} & \textbf{0.6853} {\tiny $\pm$ 0.03} \\

\bottomrule
\end{tabular}
}